\documentclass[letterpaper]{article}
\usepackage{aaai20}
\usepackage{times}
\usepackage{helvet}
\usepackage{courier}
\usepackage{amsmath}
\usepackage{amssymb}
\usepackage[ruled,vlined,linesnumbered]{algorithm2e}
\usepackage{algpseudocode}
\title{How to Evaluate Machine Learning Approaches for Combinatorial Optimization: \\
Application to the Travelling Salesman Problem}

\author{Antoine Fran\c{c}ois\textsuperscript{1,2}, Quentin Cappart\textsuperscript{1,3} \and Louis-Martin Rousseau\textsuperscript{1} \\
  \textsuperscript{1}{Ecole Polytechnique de Montr\'{e}al, Montreal, Canada} \\
  \textsuperscript{2}{Ecole des Ponts Paristech, Paris, France} \\
  \textsuperscript{3}{Element AI, Montreal, Canada} \\
  antoine.francois@eleves.enpc.fr \\
  \{quentin.cappart, louis-martin.rousseau\}@polymtl.ca}
  
\usepackage{dsfont}
\usepackage{graphicx}
\usepackage{color}
\usepackage{comment}
\usepackage{mathtools}
\usepackage{amsfonts}
\usepackage{tabularx}
\usepackage{float}
\usepackage{arydshln}
\usepackage{url}
\newtheorem{defi}{Definition}
\usepackage[lofdepth,lotdepth]{subfig}

\newtheorem{exmp}{Example}

\newcommand\qcap[1]{\textcolor{red}{#1}}

\begin{document}
\maketitle

\begin{abstract}
Combinatorial optimization is the field devoted to the study and practice of algorithms that solve NP-hard problems.
As Machine Learning (ML) and deep learning have popularized, several research groups have started to use ML to solve combinatorial optimization problems, such as the well-known \textit{Travelling Salesman Problem} (TSP).
Based on deep (reinforcement) learning, new models and architecture for the TSP have been successively developed and have gained increasing performances. At the time of writing, state-of-the-art models provide solutions to TSP instances of 100 cities that are roughly 1.33\% away from optimal solutions.  However, despite these apparently positive results, the performances remain far from those that can be achieved using a specialized \textit{search procedure}.
In this paper, we address the limitations of ML approaches for solving the TSP and investigate two fundamental questions: (1) how can we measure the level of accuracy of the pure ML component of such methods; and (2) what is the impact of a search procedure plugged inside a ML model on the performances? To answer these questions, we propose a new metric, \textit{ratio of optimal decisions} (\texttt{ROD}), based on a fair comparison with a parametrized oracle, mimicking a ML model with a controlled accuracy. 
All the experiments are carried out on four state-of-the-art ML approaches dedicated to solve the TSP. Finally, we made  \texttt{ROD} open-source in order to ease future research in the field.

\end{abstract}

\section{Introduction}

In the last few decades, Machine Learning (ML) \cite{bishop2006pattern} has progressively replaced human and expert systems to solve numerous tasks. For instance, the first algorithms in computer vision were based on hand-crafted features. Nowadays, such algorithms are learned end-to-end using Deep Learning (DL) \cite{lecun2015deep} and outperform all the traditional approaches. Similar examples are also present in speech recognition, machine translation, and in many other tasks \cite{bahdanau2014neural,chorowski2015attention,silver2017mastering}. \textit{Combinatorial Optimization Problems} (COP) \cite{parker2014discrete} have also been recently addressed by several ML based approaches.

Traditional methods dedicated to solving COPs can be classified into two main categories. The first, \textit{exact methods} (such as integer programming or constraint programming) are based on a clever exploration of a search tree and provide the optimal solution if we allow the algorithm to run completely. The drawback is their prohibitive execution cost, which makes them unsuitable for large instances. The second, \textit{heuristics}, are algorithms that are often fast to find solutions, but cannot provide any theoretical guarantees on their quality.

Although DL has also been considered with exact methods \cite{khalil2016learning,cappart2018improving,gasse2019exact}, its most popular use in combinatorial optimization is the design of heuristics. As for more traditional tasks, the \textit{holy grail} is to have a model able to learn  a heuristic end-to-end that solves a specific NP-hard problem.

The most famous and widely studied problem in combinatorial optimization is the well-known \textit{Travelling Salesman Problem} (TSP), where even its simplest version, defined on a 2D euclidean graph, has been proven to be NP-hard \cite{karp1972reducibility}. Despite the theoretical complexity of this problem, the Operations Research (OR) community has managed to build efficient algorithms for solving it \cite{applegate2006traveling}. In the last few years, the TSP has also been tackled by many DL approaches that leveraged different DL architectures and algorithms, however they remain far below the performance of the OR traditional approaches. 

By definition, ML infers knowledge from data in order to be able to transfer it to unseen similar situations. The challenge in reaching state-of-the-art performances seems to indicate that resorting only to learned knowledge is not enough to be able to have near-optimal solutions for the TSP. Most ML solutions to the TSP thus also rely on a \textit{search procedure}, characterized by a more costly execution time. A search procedure is the backbone of all combinatorial optimization algorithms (branch-and-bound, constraint programming, local search, etc.). This brings us to a fundamental question: 

\textit{What is the importance of learning versus searching in ML-based approaches to combinatorial optimization?}

This paper attempts to provide the first answer to this question by proposing a new evaluation metric, \textit{ratio of optimal decisions} (\texttt{ROD}), based on a fair comparison of learning approaches with a \textit{parametrized oracle} that is able to predict the individual decisions of a COP with a certain level of prescribed accuracy. Intuitively, a model that demonstrates a similar performance as a parametrized oracle that has poor accuracy is a sign that there is still room to improve the learning phase. Conversely, if its performance equals one of a highly accurate oracle instead, the improvements on the overall methods will most likely come from a better search procedure. 

Based on this idea, the technical contributions of this paper are as follows: 
(1) a new metric, \texttt{ROD}, for evaluating ML approaches dedicated to solve a COP that evaluates the accuracy of the learning component in isolation from the search component;
(2) the application of the metric for re-evaluating the state-of-the-art ML models for the TSP. The results show that even if the optimality gap is far worse than the traditional OR approaches, the performances of the learning component of published ML approaches are nevertheless equal to highly accurate oracles;
(3) empirical evidence that the design of the search procedure has a tremendous impact on the performance of a ML approach;
(4) the open-source release of the metric in order to help the development of future ML models for COPs.

This paper is structured as follows. The following section presents the most influential and recent developments of approaches dedicated to solve the TSP.
Next, the shortcomings of the optimality gap for evaluating ML models are described. It motivates the use of our new metric, \texttt{ROD}, which is presented thereafter. It is this particular section that is the core contribution of the paper. Finally, experiments showing the application of the metric on recent TSP models are carried out in the last section.

\section{Literature Review}

The \textit{Traveling Salesman Problem} (TSP) is a traditional COP that has been extensively studied in the literature.
Given a weighted graph, the goal is to find the shortest possible path that visits each vertex exactly once.
Finding the optimal tour is NP-hard \cite{karp1972reducibility}.
It is also true for the 2D euclidean TSP \cite{papadimitriou1977euclidean}, that considers fully connected graphs where the edges are weighted by the euclidean distances between the vertices. In practice, traditional TSP solvers 
rely on handcrafted heuristics to guide the search procedure in order to find high-quality solutions. 
Efficient approaches exist, both for exact methods and heuristics.

At the time of writing, the state-of-the-art approaches for solving TSPs are as follows. For exact methods, 
it is the well-known Concorde solver \cite{applegate2006traveling}, which is able to solve and  prove optimality to instances up to 109,399 nodes\footnote{http://www.math.uwaterloo.ca/tsp/uk/index.html} but with the prohibitive computation time of $7.5$ months. 
On the heuristic side, the most efficient approach is a variant of the Lin-Kernighan-Helsgaun algorithm (LKH) \cite{lin1973effective,helsgaun2000effective}, that has been successively refined across the years \cite{helsgaun2009general,taillard2019popmusic}. It is able to find solutions to instances of $10^7$ nodes with a duality gap of 0.584\%, according to the Held-Karp lower bound \cite{held1970traveling}. 

As machine learning has popularized, especially with the rise of deep learning \cite{lecun2015deep}, the TSP has been of particular interest for DL practitioners because they not only have the ambition to learn end-to-end new heuristics for this problem, but they are also willing to show that ML can play an important role in solving COPs. As far as we know, the TSP is the NP-hard problem that has been the most frequently considered for evaluating new ML models. It serves then as a reference, in a similar way as the MNIST dataset \cite{lecun1998gradient}, that is still used as a baseline for evaluating classification models.

The first notable ML approach dealing with the TSP was introduced by \cite{hopfield1985neural}, who solved small instances (up to 30 nodes) by the means of a Hopfield-network. More recently, new approaches resurfaced with, first, \cite{vinyals2015pointer}, that introduced the Pointer Network (PN) architecture, which is dedicated to output a permutation of an input sequence. In a case study, they apply the PN for solving euclidean TSP. It is done in a surpervised manner, and 
a \textit{beam-search} procedure is used in order to construct the final solution. Then, the PN was reused by \cite{bello2016neural}
who replaced the supervised training by reinforcement learning (RL) \cite{sutton2018reinforcement} through policy gradients methods \cite{williams1992simple} and a variant of the A3C algorithm \cite{mnih2016asynchronous}. The method is then improved with two search strategies, \textit{sampling} and \textit{active search}.
Nevertheless, \cite{deudon2018learning} proposed another encoder-decoder architecture, but enriched it with an attention mechanism \cite{vaswani2017attention}.
It is noteworthy to mention that it is also the first approach bringing the standard 2-OPT local search procedure on top of their model.

Moreover, \cite{khalil2017learning} propose to leverage another DL architecture for tackling combinatorial optimization problems over graphs, such as the TSP.
This architecture is called \texttt{structure2vec} \cite{dai2016discriminative} and is dedicated to embedding the vertices of a graph into features while keeping information on the structure of the graph. The problem is solved using neural fitted Q-learning \cite{riedmiller2005neural}.
The TSP tours are constructed step-by-step thanks to a greedy insertion method able to place each new vertex in the locally optimal position within the partially formed tour. \cite{kool2018attention} combine the ideas of a graph embedding, an encoder-decoder architecture, and a graph attention network \cite{velivckovic2017graph} with the REINFORCE RL algorithm \cite{williams1992simple} and \textit{sampling} which is then put into practice for the decoding. 

Finally, \cite{joshi2019efficient} came back to supervised learning, and make use of residual gated graph convolutional networks \cite{bresson2017residual}. 
Unlike the other approaches, the model does not output a valid TSP tour, but rather a probability for each edge of being part of the tour. The final TSP tour is computed afterwards using a \textit{greedy search} or a \textit{beam-search} procedure.

Despite the increasing performances, no ML model competes with Concorde nor the LKH algorithm. However, they are nevertheless able to find a feasible solution far more quickly by leveraging learned knowledge. Once the model has been trained, \cite{kool2018attention} reported that approximate solutions with an average optimality gap of 4.53\% can be found in 6 seconds for a data-set of 10000 euclidean TSPs of 100 nodes. For comparison, Concorde was able to solve the same instances at optimality in three minutes. By integrating a sampling selection,  \cite{kool2018attention} could reduce the optimality gap at 2.26\% but at the expense of an execution time of one hour.

Interestingly, these recent ML approaches make use of various construction techniques (beam-search, active search, sampling, 2-OPT, etc.) but without discussing how they balanced the trade-off between performance and execution time. While spending more time in the search procedure will improve the performance, it will also increase the execution time. Finding an appropriate balance between both is only mentioned by \cite{joshi2019efficient} who identify it as a future challenge. Moreover, none of these papers discuss the learning accuracy of their learning component.

That being said, the development of ML approaches for the TSP has put a strong focus on improving the learning phase. The search phase was only done by simple methods, although there exists a myriad of more refined search procedures in the literature, such as simulated annealing \cite{kirkpatrick1983optimization}, tabu search \cite{glover1998tabu}, large neighbourhood search \cite{pisinger2010large}, and many others \cite{aarts2003local}.

Based on these observations, our motivation is to provide more tools in order to help researchers make more informed choices when designing ML approaches for tackling COPs, in general.  This paper answers this question by proposing a new metric for evaluating the pure learning component of ML models. This metric is complementary to the optimality gap, which, while  commonly used, also suffers from some shortcomings.

\section{Shortcomings of the Optimality Gap Metric}

The \textit{optimality gap}, defined as the relative distance between an approximate solution and the optimal solution, is a standard and widely-used metric for evaluating approaches that solve COPs. Its main advantage is its simplicity, combined with its practical use (the performance of a model applied on a specific problem is summarized into a single value). It gives a good sense of how far we are from the optimal solution and how efficient the model is for finding the best solution. When the problem is still open and the optimal solution has not yet been proven, a relative gap can be computed using the best known solution as a baseline or using a dual bound, such as the linear relaxation or the Help-Karp lower bound for the TSP.  The focus of such metrics is put on the \textit{quality of the solution} where a solution is abstracted through a \textit{sequence of decisions}.

Generally speaking, the performance of a ML model is evaluated through a comparison with a known \textit{ground truth}, often obtained beforehand by human experts. For classification and regression tasks, this ground truth is the labelled test set that we want to access. For more complex tasks, the evaluation is done by metrics comparing the model output with references produced by human experts. For instance, the \texttt{ROUGE} metric is often used for evaluating text summarization tasks \cite{lin2004rouge}, the \texttt{BLEU} metric for translation \cite{papineni2002bleu}, or the \texttt{ELO} rating for reinforcement learning agents playing games \cite{silver2017mastering}. However, only the optimality gap has been used so far as the main evaluating metric for ML approaches tackling the TSP. Although the optimality gap also provides a comparison with a ground truth (i.e., the algorithm providing the optimal solution), we argue that this metric is not sufficient to evaluate the performance of ML approaches to COPs as it measures both the learning and search components together. If poor results are achieved, one does not really know whether the issues come from an insufficient learning ability or a weak search mechanism.

\begin{figure}[!ht]
 \centering
    \subfloat[Optimality gap of $0 \%$.]{
      \includegraphics[width=0.48\columnwidth]{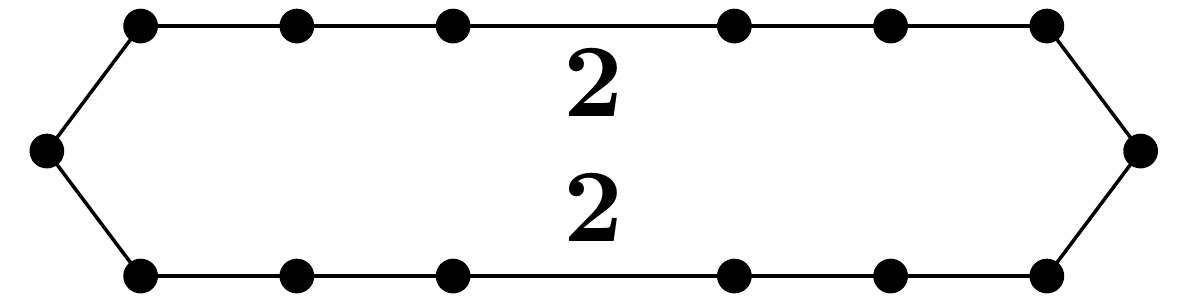}
      \label{sub:tsp-opt}
                         }
    \subfloat[Optimality gap of $27.3 \%$.]{
      \includegraphics[width=0.48\columnwidth]{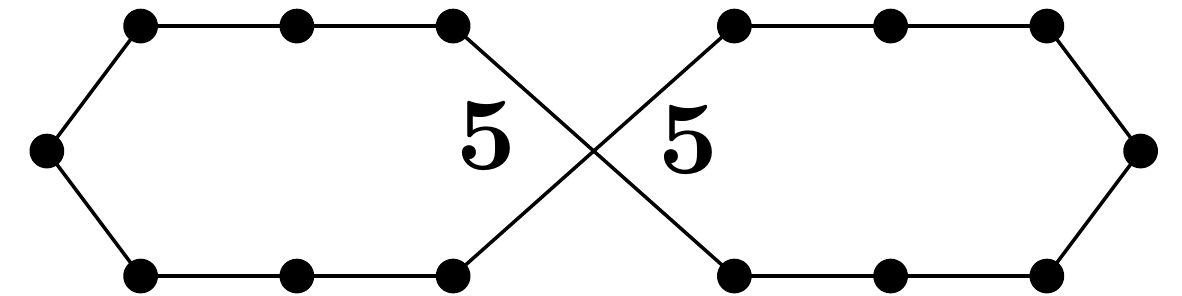}
      \label{sub:tsp-sub}
                         } 
    \caption{Pathological example of a bad decision in the TSP.}
    \label{fig:tsp}
\end{figure}

This problem is illustrated in Fig.~\ref{fig:tsp}. The left figure presents the optimal solution of a TSP (non labelled edges have a weight of 1). On the right, a sub-optimal solution is proposed. Although only one non-optimal decision has been done (the second misplaced edge being forced to complete the tour), the optimality gap is huge. When considering only the optimality gap, one might believe that a new DL architecture could help. However, the quasi-totality of edges were guessed correctly and a simple search heuristic would have fixed the problem. For this reason, we advocate the use of a second metric for evaluating ML models dedicated to solve COPs. The idea is to have a metric able to compare the accuracy of the learning component against a ground truth, independently of the search component.


\section{ROD: Evaluating only the learning component}

This section describes \textit{ratio of optimal decisions} (\texttt{ROD}), a new metric we introduce for evaluating the learning component of ML approaches dedicated to solve COPs. Its focus is not on the quality of the solution, such as the optimality gap, but rather on the \textit{quality of the individual decisions}, which better reflects how good a model is performing against a ground truth.

When solving a COP, one has to assign a specific value to a set of variables in order to find a feasible assignment that minimizes (resp. maximizes) an objective function. A simple and general way to model such a problem is to use a \textit{Dynamic Programming} (DP) formulation. The idea is to simplify the problem by breaking it down into a sequence of decision steps. At each step, a new variable is selected and assigned a value until all the variables have been set. A cost is induced after each assignment and the total cost, when all decisions are taken, corresponds to the outcome of the objective function. A fundamental property of DP is the so-called \textit{principle of optimality}, introduced by \cite{bellman1966dynamic}: a sequence of optimal decisions done at each step gives the optimal solution of the complete problem.

In the ML terminology, an optimal oracle can be considered as a model having a \textit{perfect knowledge}, which means that it never takes sub-optimal decisions. Let us assume that we have a \textit{parametrized oracle} that is able to take each optimal decision with a certain accuracy. Based on this idea, the \texttt{ROD} metric we introduce is defined as follows.



\begin{defi}[Ratio of optimal decisions (ROD)]
Let $P$ be a COP, $M$ be a model dedicated to solve $P$, and $\Omega$ be a parametrized oracle that is able to take each optimal decision with a certain accuracy.
The \texttt{ROD} of $M$ regarding $\Omega$ is defined as the ratio of optimal decisions that is required by $\Omega$ in order to equal the optimality gap of $M$ on $P$. 
\end{defi} 

\begin{exmp}
Let us consider the TSP of Fig.~\ref{fig:tsp} that contains 13 decisions, and let us assume that the solution of Fig.~\ref{sub:tsp-sub} has been obtained using a parametrized oracle $\Omega$.
Models having an optimality gap of 27.3\% will have a \texttt{ROD} of $12/13 = 92.31\%$ regarding $\Omega$ because they will have the same performance as a parametrized oracle that has made 12 optimal decisions among 13.
\end{exmp}

The goal of \texttt{ROD} is to measure the break-even ratio of a parametrized oracle, where its performance equals the one of the model we want to evaluate. 
This
ratio will indicate that the model has similar performances as a function of having a parametrized level of knowledge.
By doing so, the model is directly compared with a ground truth only, without integrating the difficulty of the problem. 
Indeed, if both of them make the same bad decision, the optimality gap increase will remain the same for both.

The last question remaining is how \texttt{ROD} can be computed.
This question is tightly related to the construction of the parametrized oracle.
Two design choices are required for that: 
(1) given a ratio, how do we select the decisions that will be optimal;
and (2) when a non-optimal decision is made, which one must be selected?
These questions are discussed in the next section.

\section{Construction of the Parametrized Oracle}

The formulation proposed is generic, in the sense that it can be used for any COP. Let us consider a COP $P = \langle X,D,C,O \rangle$ where $X$ is the set of variables, $D$ the set of domains restricting the values that variables $x \in X$ can take, $C$ the set of constraints and $O$ the objective function. 

The first step is to design a DP model associated to the COP. It requires adequately defining the tuple $\langle S,A,T, K \rangle$ where $S$ is the set of all the possible states that can be generated,
$A$ is the set of possible actions, $T: S \times A \to S$ is the transition function leading the system to move from a state to another one given the action taken and $K : S \times A \to \mathbb{R}$ is the cost function returning the cost of every action taken in a given state.
The DP model of the parametrized oracle is defined as follows.

\begin{description}

\item[State] Let $n$ be the number of variables $x\in X$ and $\sigma=\langle x_1, \dots, x_n \rangle$ be an arbitrarily ordered sequence of these variables.
A state $s \in S$ corresponds to the first non-assigned variable $x_i \in \sigma$, with $i \in \{1,\dots,n\}$, or $\bot$ if all of them are assigned. The initial state is then $x_1$ and a state is terminal when $s = \bot$.

\item[Action] An action $a_s$ done at state $s$ is defined as the selection of a value for its assignment to the variable referenced by $s$. The action is valid if and only if (1) it is in the domain of the variable ($a_s \in D(s)$), and (2) its assignment does not violate any constraint $c \in C$.


\item[Transition] The transition function is defined as $T(s, a_s) = s \leftarrow a_s$, where $x \leftarrow a$ is a function assigning the value $a$ to the variable $x$ and returning the next non-assigned variable from the sequence $\sigma$.

\item[Cost Function] The cost function $K(s, a_s)$ corresponds to the increase on the objective cost that is caused by the assignment of the value $a_s$ to the variable associated to $s$.

\end{description}

Once defined, the DP model has to be solved by means of a \textit{policy}. For each state, we must decide what action to perform.
To do so, we make use of the function  $\Theta: S \times [0,1] \to A$, which corresponds to the decision suggested by the parametrized oracle introduced in the previous section. 

This function predicts the optimal action $a_s^\star$ for a state $s$ with a probability $\alpha$, otherwise it selects a \emph{relatively good} action by sampling it randomly from the set of the remaining actions. The sampling is done using a weighted distribution $\mathbb{P}$ that favors actions generating low cost increases. By doing so, the parametrized oracle's  behaviour will be more similar to ML models. If a sub-optimal decision is suggested, it is more likely that the model will choose a good action instead of a poor one.
This weighted distribution is defined as follows.


\begin{equation}
\label{def:proba_distribution}
\mathbb{P}(a) =  \frac{K(s,a)^{-1}}{ \sum_{a^\prime \in D(s)} K(s,a^\prime)^{-1}} \ \ \ \ \ \forall s \in S, \forall a \in D(s)
\end{equation}

The negative exponent is for weighting the actions by their inverse cost.
Then, $\Theta$ is defined in Eq. \eqref{def:oracle}, where $\thicksim_\mathbb{U}$ denotes a uniform sample from a set, and $\thicksim_\mathbb{P}$ a sample following the distribution of Eq. \eqref{def:proba_distribution}.

\begin{equation}
\label{def:oracle}
 \Theta(s, \alpha) = 
\begin{cases}
    a^\star_s  \in  D(s)  & \text{if} \ ÷  \epsilon \thicksim_\mathbb{U} [0,1] \leq \alpha. \\
    a _s \thicksim_\mathbb{P}  D(s) & \text{otherwise}.
\end{cases}
\end{equation}

Finally, the complete oracle $\Omega(P,\Theta, \alpha)$ is a function taking as input a COP $P$ and a policy $\Theta$ parametrized by $\alpha$ and which outputs the cost of a solution of $P$ according to $\Theta$.
The \texttt{ROD} of a ML model $M(P)$ we want to evaluate on $P$, corresponds to the probability $\alpha$ yielding an oracle having the same optimality gap as $M$, on average. In practice, one can also compute \texttt{ROD} of a model evaluated on many instances of the same COP.

An important note is that the parametrized oracle does not resort to any kind of search procedure: no look-aheads are allowed and decisions cannot be undone.
This design choice is made in order to have a fair comparison with a pure ML model that does not use local search corrections.

Using the parametrized oracle, \texttt{ROD} is computed by increasing $\alpha$ step-by-step until $\Omega$ and $M$ returns the same optimality gap for a data set of instances we want to evaluate.
The computation of  \texttt{ROD} is shown in Alg.~\ref{alg:ROD}. 
As long as the current oracle $\Omega$ has a lower performance than $M$, the ratio is increased, otherwise it is returned and it
corresponds to the  \texttt{ROD} value. The exact optimality gap can be obtained using the perfect oracle ($\alpha = 1$) for computing the optimal solution of each instance.

\algrenewcomment[1]{\(\triangleright\) #1}

\begin{algorithm}[!ht]

\Comment{\textbf{Pre:} $\mathcal{M}$ is the ML model we want to evaluate.}

\Comment{\hspace{0.7cm} $\mathcal{D}$ is the test set containing COP instances.}

\Comment{\hspace{0.7cm} $\Omega$ the parametrized oracle.}

\Comment{\hspace{0.7cm} $\Theta$ is the oracle function as defined previously.}

\Comment{\hspace{0.7cm} $k$ is the step size for increasing $\alpha$.}

$\alpha := 0$

\While{$\alpha \leq 1$}{

    $c^\star := 0, \ c^\alpha := 0, \ c^M := 0$
    
    \For{$P \in \mathcal{D}$}{
    
           $c^\star := c^\star + \Omega(P,\Theta,1)$     \hfill \Comment{\textit{Optimal model}}
           
           $c^\alpha := c^\alpha + \Omega(P,\Theta,\alpha)$     \hfill \Comment{$\alpha$-\textit{oracle}}
           
           $c^M := c^M + M(P)$     \hfill \Comment{\textit{ML model}}
          
    }
    
    $c^\alpha := 1 - \frac{c^\star}{c^\alpha}$ \hfill \Comment{\textit{Average optimality gap}}
    
    $c^M := 1 - \frac{c^\star}{C^M}$

    \If{$ c^\alpha \leq c^M$}{
        \textbf{break} \hfill \Comment{$\alpha$ \textit{is the} \texttt{ROD} value}
    }
    
    $\alpha := \alpha + k$

}

\Return $\alpha$ 
\caption{Computation of \texttt{ROD}.}

\label{alg:ROD}
\end{algorithm}

\section{Case Study: the Travelling Salesman Problem}

Two sets of experiments are carried out. First, we use \texttt{ROD} to re-evaluate state-of-the-art ML models for solving TSP on 2D euclidean graphs. Then, we propose and analyze several standard search procedures that can be used in order to improve the performances for each model.

\subsection{Problem Definition}

\begin{defi}[Travelling Salesman Problem (TSP)]
Let $G=(V,E)$ be a simple weighted graph of $n$ vertices. A tour $(x_i)_{1 \le i \le n}$ of  $G$ is a permutation of the vertices
such that it is possible to go through every vertex ($V$) exactly once by following the edges ($E$) and come back to the initial vertex. The cost $c$ associated with a tour corresponds to the sum of the weights of edges that are followed.
The TSP consists in finding a tour with minimal cost.
\end{defi}

With the 2D euclidean variant, the graph is fully connected and the edges are weighted by the euclidean distances between the vertices. The TSP can be formalized as a COP $\langle X,D,C,O\rangle$, as follows. Each variable $x_i \in X$ corresponds to a vertex $v_i \in V$ of the graph and indicates what will be the next vertex to visit after leaving $x_i$. We then have $D(x_i) = V \backslash \{v_i\}$. The only constraint is that the final tour must be a permutation (i.e. a circuit) of the set of vertices. In the literature, it is often expressed through the circuit constraint: $\texttt{circuit}(X)$ \cite{beldiceanu1994introducing}. Finally, the objective function is to minimize the cost $c$ of the tour.
As described in the previous section, a COP has an associated DP formulation $\langle S,A,T, K \rangle$, which corresponds to the parameterized oracle.

\subsection{ML Models and Search Procedures Considered}

The models we selected are summarized in Table \ref{tab:usedpapers}.
While all of them are dedicated to the TSP or close variants, they differ by their neural architecture and the learning algorithm considered. Graph Convolutional Networks (GCN) \cite{dai2016discriminative,bresson2017residual} and Graph Attention Networks (GAT) \cite{velivckovic2017graph} are two standard and popular architectures when dealing with problems having a graph structure. 
In regards to the training, the trend is more on RL approaches that either use Deep Q-learning (DQN) or policy gradient methods like REINFORCE; although supervised training is also considered.  

Different search procedures have been also considered for improving the solution obtained by the model. 
Many neural networks output probability distributions that give insight on what edges should be used to construct an optimal solution.
\textit{Greedy decoding} consists in taking the edges having the highest probability, whereas \textit{sampling} consists in selecting some of them randomly, based on the distribution that has been inferred. The best solution found is then returned. \textit{Beam-search} is a different construction approach where, at each step, we keep only the $b$ best partial solutions, and where $b$ is a parameter referred to as the \textit{the beam-width}.
This procedure has been improved by  \cite{joshi2019efficient} and uses a shortest path heuristic for closing the TSP tour.

On top of that, one can also integrate perturbative search heuristics that modify a complete solution in order to improve it. Famous examples are the 2-OPT, 3-OPT and Lin-Kernighan heuristic (LK) \cite{lin1973effective} 
that swap edges of the current solution to reduce the tour cost. Perturbative search procedures have been less considered in the previous ML models, with the exception of \cite{deudon2018learning} that use the 2-OPT heuristic.

\begin{table}[!ht]
\caption{Models considered for evaluation with \texttt{ROD}.}

\centering
\resizebox{\columnwidth}{!}{
\begin{tabular}{|l|l|l|l|}
\hline
Approach & Model & Learning  & Search \\
\hline
\hline
Khalil et al. 2017 & GCN & DQN & Greedy \\
\hline
Deudon et al. 2018  & GAT & REINFORCE & Sampling \\
  &  &  & 2-OPT \\
\hline
Kool et al. 2018 & GAT &  REINFORCE & Greedy  \\
  &  &  & Sampling \\
\hline
Joshi et al. 2019 & GCN & Supervised & Greedy \\
  &  &  & Beamsearch \\
 & &  & Shortest tour \\
\hline
\end{tabular}
}
\label{tab:usedpapers}
\end{table}


\subsection{Experimental Protocol}

In order to analyze the models of Table~\ref{tab:usedpapers},
we resort to the source code the authors provided with the related paper
\footnote{https://github.com/Hanjun-Dai/graph\_comb\_opt \\
https://github.com/MichelDeudon/encode-attend-navigate\\
https://github.com/wouterkool/attention-learn-to-route \\
https://github.com/chaitjo/graph-convnet-tsp}. 
For \cite{kool2018attention,joshi2019efficient}, the models had already been trained by the authors, so we used them as is. Otherwise, we retrain them using the procedure described.
At first, as we are interested in analyzing the quality of the learned model, we modified the codes in order to only allow a greedy decoding. 

For the evaluation, we implemented the \texttt{ROD} construction as presented in Alg.~\ref{alg:ROD}.
Two test sets of 1000 random 2D euclidean graphs, with 50 and 100 vertices, are considered. 
The instances are generated by uniformly sampling the vertices on the unit 2D square. All the vertices are then connected using the euclidean norm for weighting the edges. The optimal tours of the instances are computed using the Concorde solver \cite{applegate2006traveling}. 
Concerning the search procedures, we reused an open-source implementation of the 2-OPT, 3-OPT, and LK heuristic\footnote{https://gitlab.com/Soha/local-tsp} and the search mechanisms already integrated in the ML models as well.


For the reproducibility of results and to further help  research in this field, the implementation of \texttt{ROD} is available online\footnote{https://github.com/qcappart/ROD\_oracle}. Finally, for the neural network computations, all the experiments have been carried out on a single Tesla V100 PCIe 32GB GPU, while the rest of the operations were done on Intel Xeon Silver 4116 CPUs.

\subsection{Application of ROD}

\begin{table*}[ht]
    \renewcommand{\arraystretch}{1.1}
\centering
\begin{tabular}{|l|rrr|rrr||rrr|rrr|}
\hline
  & \multicolumn{6}{c||}{\textbf{Evaluation on TSPs of 50 vertices}}    & \multicolumn{6}{c|}{\textbf{Evaluation on TSPs of 100 vertices}} \\ 
  \hline
Number of  & \multicolumn{3}{c|}{optimality gap (\%)} & \multicolumn{3}{c||}{\texttt{ROD} (\%)}  & \multicolumn{3}{c|}{optimality gap (\%)} & \multicolumn{3}{c|}{\texttt{ROD} (\%)} \\ 

vertices for training & \multicolumn{1}{c}{20} & \multicolumn{1}{c}{50} & \multicolumn{1}{c|}{100} & \multicolumn{1}{c}{20} & \multicolumn{1}{c}{50} & \multicolumn{1}{c||}{100} 
& \multicolumn{1}{c}{20} & \multicolumn{1}{c}{50} & \multicolumn{1}{c|}{100} & \multicolumn{1}{c}{20} & \multicolumn{1}{c}{50} & \multicolumn{1}{c|}{100} \\ 
\hline
Khalil et al. (2016) & 11.01 & 9.29 & 8.74 & 96.4 & 97.1 & 97.3 & \textbf{10.66} & 10.27 & 10.74 & \textbf{97.9} & 98.0 & 97.9 \\
Deudon et al. (2018) & 9.18 & 4.87 & 7.46 & 97.1 & 98.3 & 97.6 & 23.03 & 9.03 & 8.38 & 95.4 & 98.3 & 98.4 \\
Kool et al. (2018) & \textbf{4.35} & \textbf{1.69} & \textbf{4.22} & \textbf{98.5} & \textbf{99.5} & \textbf{98.6} & 16.11 & \textbf{4.89} & \textbf{4.35} & 96.8 & \textbf{99.1} & \textbf{99.2} \\
Joshi et al. (2019)& 42.58 & 4.41 & 38.85 & 85.0 & 98.6 & 86.5 & 70.64 & 52.92 & 8.61 & 83.9 & 88.2 & 98.3 \\
\hline
Concorde Solver & 0 & 0 & 0 & - & - & - & 0 & 0 & 0 & - & - & - \\
\hline
\end{tabular}
\caption{Application of the optimality gap and \texttt{ROD} on the learning component of ML models using different configurations.}
\label{tab:oracleprecisions}
\end{table*}


Table \ref{tab:oracleprecisions} reports the optimality gap and \texttt{ROD} on the two test sets for the learning component of the ML models previously described. Three configurations for the training are considered (instances of 20, 50, and 100 vertices).  First of all, we notice that the optimality gap obtained is consistent with the results published by the authors. When considering only the learning component of the models, 
\cite{kool2018attention} is the most efficient. It is also noteworthy to mention that \texttt{ROD} metric is consistent with the optimality gap. An increase of the former is characterised by a decrease of the latter.
Standard analysis that were done using the optimality gap as a baseline can still be performed with \texttt{ROD}. For instance, the approach of \cite{joshi2019efficient} seems to suffer from over-fitting: the performances decrease drastically when the training and the evaluation are done with instances of a different size.

In any case, \texttt{ROD} provides new information that remained hidden with the optimality gap metric. 
First, we observe that in all the situations the optimality gap is far from what is achieved by Concorde, which provides optimal solutions.
Initially, one could think that there is still room for improving the learning component of these models. 
However, \texttt{ROD} indicates that most of the models already achieve high-quality performances by equaling parametrized oracles taking, on average, more than 95\% of the optimal decisions during the construction of the solution.
As a result of these high performances, designing new competitive learning components would not be a trivial task, and improvements may come from a search procedure instead.

Following the same idea, we can also notice that \texttt{ROD} remains stable when evaluation is done on instances on 100 vertices instead of 50. 
For \cite{kool2018attention}, the difference is less than 2\% in the worst case, which indicates a relatively good generalization. Contrarily, the related optimality gap, when trained on 20 nodes, have an increase exceeding 10\%. 
This observation gives another indication that the optimality gap is not suited for evaluating the learning ability of ML models for COPs.

\subsection{Impact of the Search Procedures}

The previous set of experiments showed that further improvements may not come from a better learning ability but, rather, from a search procedure instead. The goal of this second set of experiments is to analyse 
the impact of different standard search procedures performed during the evaluation. To do so, the greedy construction, sampling and beam-search are considered for constructing the first solution from the learned model, and 2-OPT, 3-OPT, and the Lin-Kernighan heuristic (LK) are added to improve it. The impact of these search procedures is analyzed in Table \ref{2OPT_opt_gaps}. The optimality gap of the ML models from Table \ref{tab:usedpapers}, with the improvements done by the search procedures, are reported. The best results among the ML models are highlighted.
The training and the evaluation are done on TSPs of 100 vertices and the test set contains 1000 instances. 16 iterations are considered when sampling, as well as a beam-width of 16 for the beam-search (BS). For the special case of \cite{joshi2019efficient}, their shortest-path heuristic (BS*) is also analyzed. The optimality gap, the impact of the search procedure on the optimality gap ($\Delta$: the reduction of the optimality gap from the situation without the search procedure), and the average execution time are reported. As a baseline, the performances of the \textit{nearest neighbour} (NN) heuristic, that selects the closest available city in the current TSP tour, is also reported. 

At first glance, we can observe that all the methods benefit from a search procedure, both when constructing the solution (greedy, sampling, and beam-search) and when improving it (2-OPT, 3-OPT, and LK). The approaches of \cite{khalil2017learning} and \cite{joshi2019efficient} seem to be the ones that benefit the most from a search procedure ($\Delta$ value). For instance, 
the optimality gap of the greedy variant of \cite{joshi2019efficient} can be reduced from 8.61\% to 0.24\% by adding beam search and LK. However, resorting to more elaborate search procedures can only be done at the expense of a more prohibitive execution time. On this last example, the execution time is roughly 18 times greater (from 91ms to 1.7s).
There is, then, a trade-off to define in calculating the difference in performance and execution time when resorting to a search procedure. 
In practice, one can also run a search procedure until the available time has been exceeded and then return the best solution found.

Interestingly, even if the approach of \cite{kool2018attention} had the best performance when no search procedure is considered, it is then beaten by \cite{joshi2019efficient} when search is integrated. 
It empirically shows that the models do not behave in the same way with a similar search procedure. 
Finally, it is noteworthy to mention, only by plugging ML models with standard search procedures from the literature, we can achieve state-of-the-art results by reducing the optimality gap from $7.05\%$ to $0.24\%$ when a beam-width of 16 and LK heuristic are considered with the approach of \cite{joshi2019efficient}. This advocates the use of hybrid methods, combining learning and searching, when dealing with NP-hard problems.

\begin{table*}[ht]
\centering
\resizebox{\textwidth}{!}{%
\begin{tabular}{|l|l|rr|rrr|rrr|rrr|}
\hline
\textbf{Algorithm} & \textbf{Construction} & \multicolumn{2}{c|}{\textbf{Without local search}} & \multicolumn{3}{c|}{\textbf{2-OPT}} & \multicolumn{3}{c|}{\textbf{3-OPT}} & \multicolumn{3}{c|}{\textbf{LK}} \\
& & gap (\%) & time (s) &  $\Delta$ (\%) & gap (\%) & time (s) &  $\Delta$ (\%) & gap (\%) &  time (s)  & $\Delta$ (\%) &  gap (\%) & time (s) \\
\hline
\hline
NN & - & 14.82 & 0.035 &  -9.30 & 5.52 & 0.063 &  -12.39 & 2.43 & 83.0 &  -12.97 & 1.85 & 29.4 \\
\hline
\hline
Khalil et al. & Greedy & 10.74 & 0.075 & -2.74 & 8.00 &  0.093 & \textbf{-7.86} & 2.88 &  90.2 & \textbf{-9.43} & 1.31 &  23.8 \\
\hline
Deudon et al. & Greedy & 8.38 & 0.036 & -2.78 & 5.60 & 0.840 &  -4.98 & 3.40 & 70.3 &  -6.32 & 2.06 & 21.4 \\
 & S (16 it.) & 7.48 & 0.046 &  -3.22 & 4.26 & 1.196  & -4.41 & 3.07 & 79.1 &  -5.40 & 2.08 & 29.0 \\
\hline
Kool et al. & Greedy & 4.35 & 0.023 &  -0.98 & 3.37 & 0.046 &  -2.31 & 2.04 & 38.6 &  -2.89 & 1.46 & 2.5 \\
& S (16 it.) & \textbf{3.18} & 0.205 &  -0.68 & 2.50 & 0.167 & -1.58 & \textbf{1.60}  & 32.9 &  -1.93 & 1.25 & 33.1\\
\hline
Joshi et al. & Greedy & 8.61 & 0.091 & \textbf{-6.87} & 1.74 & 0.063 &  -4.98 & 3.63 & 25.6 &  -8.24 & 0.37 & 2.2 \\
 & BS ($b = 16$) & 7.05 & 0.095 &  -5.80 & 1.25 & 0.058 &  -4.38 & 2.67 & 21.8 &  -6.81  & \textbf{0.24} & 1.7 \\
 & BS* ($b = 16$) & 6.35 & 0.151  & -5.15 &  \textbf{1.20}& 0.068 &  -4.57 & 1.78 & 21.4 & -6.09 & 0.26 &  1.8 \\
\hline
\end{tabular}}
\caption{Optimality gaps obtained before/after resorting to search procedures, on a test set of 1000 instances (100 vertices).}
\label{2OPT_opt_gaps}
\end{table*}

\section{Conclusion}

A search procedure is the backbone of traditional algorithms dealing with Combinatorial Optimization Problems (COP). More recently a new paradigm, based on learning, has been considered for solving COPs.
Despite the apparently good results of learning approaches, the performances are still far from those that can be achieved when using algorithms based on a specialized search procedure.  
This observation indicates that resorting only to learned knowledge may not be enough  to  be  able  to  have  near-optimal  solutions when solving COPs. 
For this reason, state-of-the-art ML approaches also rely on a search procedure, characterized by a more costly execution time. 
Which brings us back to the question:
\textit{What is the importance of learning versus searching when designing ML approaches for solving COPs?}

The goal of this paper was to provide the first answers to this question. To do so, we target the well-known Travelling Salesman Problem, that is, to the best of our knowledge, the NP-hard problem that has been the most studied by learning approaches for COPs.

Firstly, we introduced a new evaluation metric, \texttt{ROD}, that can be used for evaluating the learning component of any ML approach dedicated to solve COPs. 
We applied this metric on the four state-of-the-art ML models from the literature and the results showed that, even if the models do no compete with traditional approaches, the learning ability of the models are nevertheless of high quality. This result suggests that further improvements may come from the search procedure, and not from the learning component itself. Future works on the TSP should then be dedicated to improving the search procedure.
In order to ease such new developments, we made \texttt{ROD} open-source.

Secondly, we experimentally showed that the four ML approaches that we tested benefited from a search procedure. Only by combining ML approaches with standard search procedures, the optimality gap obtained by \cite{joshi2019efficient} could be reduced from 7.05\% to 0.24\% for TSPs of 100 vertices. However, it is done at the expense of a higher execution time. 
Finding the most adapted search procedures for a ML model is still an open question.

So far, we used \texttt{ROD} only on the TSP. In our future work, we plan to consider it for evaluating ML models dedicated to solve other COPs, such as the Knapsack \cite{bello2016neural} or the Maximum Independent Set Problem \cite{khalil2017learning}. Moreover, we also plan to consider more elaborate search procedures (simulated annealing, tabu search, large neighbourhood search, etc.) for the evaluation.

%
%
\bibliography{paperbib}
\bibliographystyle{aaai}

\end{document}